# Polynomial Constraints in Causal Bayesian Networks

**Changsung Kang**
Department of Computer Science
Iowa State University
Ames, IA 50011
*cskang@iastate.edu*

**Jin Tian**
Department of Computer Science
Iowa State University
Ames, IA 50011
*jtian@cs.iastate.edu*

## Abstract

We use the implicitization procedure to generate polynomial equality constraints on the set of distributions induced by local interventions on variables governed by a causal Bayesian network with hidden variables. We show how we may reduce the complexity of the implicitization problem and make the problem tractable in certain causal Bayesian networks. We also show some preliminary results on the algebraic structure of polynomial constraints. The results have applications in distinguishing between causal models and in testing causal models with combined observational and experimental data.

## 1 Introduction

The use of graphical models for encoding distributional and causal information is now fairly standard [Heckerman and Shachter, 1995, Lauritzen, 2000, Pearl, 2000, Spirtes *et al.*, 2001]. The most common such representation involves a *causal Bayesian network (BN)*, namely, a directed acyclic graph (DAG) $G$ which, in addition to the usual conditional independence interpretation, is also given a causal interpretation. This additional feature permits one to infer the effects of interventions or actions, such as those encountered in policy analysis, treatment management, or planning. Specifically, if an external intervention fixes any set $T$ of variables to some constants $t$, the DAG permits us to infer the resulting post-intervention distribution, denoted by $P_t(v)$,[1] from the pre-intervention distribution $P(v)$. The quantity $P_t(y)$, often called the "causal effect" of $T$ on $Y$, is what we normally assess in a controlled experiment with $T$ randomized, in which the distribution of $Y$ is estimated for each level $t$ of $T$. We will call a post-intervention distribution an interventional distribution, and call the distribution $P(v)$ non-experimental distribution.

The validity of a causal model can be tested only if it has empirical implications, that is, it must impose constraints on the statistics of the data collected. A causal BN not only imposes constraints on the non-experimental distribution but also on the interventional distributions that can be induced by the network. Therefore a causal BN can be tested and falsified by two types of data, observational, which are passively observed, and experimental, which are produced by manipulating (randomly) some variables and observing the states of other variables. The ability to use a mixture of observational and experimental data will greatly increase our power of causal reasoning and learning.

There has been much research on identifying constraints on the non-experimental distributions implied by a BN with hidden variables [Verma and Pearl, 1990, Robins and Wasserman, 1997, Desjardins, 1999, Spirtes *et al.*, 2001, Tian and Pearl, 2002]. In algebraic methods, BNs are defined parametrically by a polynomial mapping from a set of parameters to a set of distributions. The distributions compatible with a BN correspond to a *semi-algebraic set*, which can be described with a finite number of polynomial equalities and inequalities. In principle, these polynomial equalities and inequalities can be derived by the quantifier elimination method presented in [Geiger and Meek, 1999]. However, due to high computational demand (doubly exponential in the number of probabilistic parameters), in practice, quantifier elimination is limited to models with few number of probabilistic parameters. [Geiger and Meek, 1998, Garcia, 2004, Garcia *et al.*, 2005] used a procedure called *implicitization* to generate independence and non-independence constraints on the observed non-experimental distributions. These constraints consist of a set of polynomial equalities that define the smallest *algebraic set* that contains the semi-algebraic set. [Garcia *et al.*, 2005] analyzed the algebraic structure of constraints for a class of small BNs.

Algebraic approaches have been applied in causal BNs to deal with the problem of the identifiabil-ity of causal effects [Riccomagno and Smith, 2003, Riccomagno and Smith, 2004]. However, to the best of our knowledge, the implicitization method has not been applied to the problem of identifying constraints on interventional distributions induced by causal BNs.

---
[1] [Pearl, 1995, Pearl, 2000] used the notation $P(v|set(t))$, $P(v|do(t))$, or $P(v|\hat{t})$ for the post-intervention distribution, while [Lauritzen, 2000] used $P(v\|t)$.

In this paper, we seek the constraints imposed by a causal BN on both nonexperimental and interventional distributions. When all variables are observed, a complete characterization of constraints on interventional distributions imposed by a given causal BN has been given in [Pearl, 2000, pp.23-4]. In a causal BN containing hidden variables, a class of equality and inequality constraints on interventional distributions are given in [Kang and Tian, 2006]. In this paper, we propose to use the implicitization procedure to generate polynomial constraints on interventional distributions induced by a causal BN with hidden variables. The main challenges in applying the implicitization procedure on interventional distributions are:

(i) *Computational complexity*. The generic complexity of implicitization is known to be exponential in the number of variables (number of parameters for this problem). When we consider interventional distributions, the number of variables greatly increases compared to the case of non-experimental distribution, which makes the computation infeasible even for small causal BNs.

(ii) *Understanding structures of constraints*. Finding a syntactic structure of the constraints computed by implicitization also becomes complicated.

To deal with challenge (i), we show two methods to reduce the complexity of the implicitization problem. We illustrate our method showing a model in which the generic implicitization procedure is intractable while our methods can solve the problem. We also show an example of new constraints on interventional distributions that are not captured by the types of constraints in [Kang and Tian, 2006]. To deal with challenge (ii), we present some preliminary results on the algebraic structure of polynomial constraints on interventional distributions implied by certain classes of causal BNs with hidden variables. We also present some preliminary results in causal BNs without hidden variables, which are often useful in understanding syntactic structures of the constraints for BNs with hidden variables.

## 2 Preliminaries and Problem Statement

### 2.1 Causal Bayesian Networks and Interventions

A causal Bayesian network, also known as a *Markovian model*, consists of two mathematical objects: (i) a DAG $G$, called a *causal graph*, over a set $V = \{V_1, \ldots, V_n\}$ of vertices, and (ii) a probability distribution $P(v)$, over the set $V$ of discrete variables that correspond to the vertices in $G$.[2] In this paper, we will assume a topological ordering $V_1 > \ldots > V_n$ in $G$. $V_1$ is always a sink and $V_n$ is always a source. The interpretation of such a graph has two components, probabilistic and causal. The probabilistic interpretation views $G$ as representing conditional independence restrictions on $P$: Each variable is independent of all its non-descendants given its direct parents in the graph. These restrictions imply that the joint probability function $P(v) = P(v_1, \ldots, v_n)$ factorizes according to the product

$$P(v) = \prod_i P(v_i|pa_i) \quad (1)$$

where $pa_i$ are (values of) the parents of variable $V_i$ in $G$.

The causal interpretation views the arrows in $G$ as representing causal influences between the corresponding variables. In this interpretation, the factorization of (1) still holds, but the factors are further assumed to represent autonomous data-generation processes, that is, each conditional probability $P(v_i|pa_i)$ represents a stochastic process by which the values of $V_i$ are assigned in response to the values $pa_i$ (previously chosen for $V_i$'s parents), and the stochastic variation of this assignment is assumed independent of the variations in all other assignments in the model. Moreover, each assignment process remains invariant to possible changes in the assignment processes that govern other variables in the system. This modularity assumption enables us to predict the effects of interventions, whenever interventions are described as specific modifications of some factors in the product of (1). The simplest such intervention, called *atomic*, involves fixing a set $T$ of variables to some constants $T = t$, which yields the post-intervention distribution

$$P_t(v) = \begin{cases} \prod_{\{i|V_i \notin T\}} P(v_i|pa_i) & v \text{ consistent with } t. \\ 0 & v \text{ inconsistent with } t. \end{cases} \quad (2)$$

Eq. (2) represents a truncated factorization of (1), with factors corresponding to the manipulated variables removed. This truncation follows immediately from (1) since, assuming modularity, the post-intervention probabilities $P(v_i|pa_i)$ corresponding to variables in $T$ are either 1 or 0, while those corresponding to unmanipulated variables remain unaltered. If $T$ stands for a set of treatment variables and $Y$ for an outcome variable in $V \setminus T$, then Eq. (2) permits us to calculate the probability $P_t(y)$ that event $Y = y$ would occur if treatment condition $T = t$ were enforced uniformly over the population.

When some variables in a Markovian model are unobserved, the probability distribution over the observed variables may no longer be decomposed as in Eq. (1). Let $V = \{V_1, \ldots, V_n\}$ and $U = \{U_1, \ldots, U_{n'}\}$ stand for the sets of observed and unobserved variables respectively. If no $U$ variable is a descendant of any $V$ variable, then the corresponding model is called a *semi-Markovian model*. In this paper, we only consider semi-Markovian models. However, the results can be generalized to models with arbitrary unobserved variables as shown in [Tian and Pearl, 2002]. In a semi-Markovian model, the observed probability distribution, $P(v)$, becomes a mixture of products:

$$P(v) = \sum_u \prod_i P(v_i|pa_i, u^i)P(u) \tag{3}$$

where $PA_i$ and $U^i$ stand for the sets of the observed and unobserved parents of $V_i$, and the summation ranges over all the $U$ variables. The post-intervention distribution, likewise, will be given as a mixture of truncated products

$$P_t(v) = \begin{cases} \sum_u \prod_{\{i|V_i \notin T\}} P(v_i|pa_i, u^i)P(u) & v \text{ consistent with } t \\ 0 & v \text{ inconsistent with } t \end{cases} \tag{4}$$

---

[2]We only consider discrete random variables in this paper.

Assuming that $v$ is consistent with $t$, we can write

$$P_t(v) = P_t(v \setminus t) \tag{5}$$

In the rest of the paper, we will use $P_t(v)$ and $P_t(v \setminus t)$ interchangeably, always assuming $v$ being consistent with $t$.

## 2.2 Algebraic Sets, Semi-algebraic Sets and Ideals

We briefly introduce some concepts related to algebraic geometry that will be used in this paper.

The set of all polynomials in $x_1, \ldots, x_n$ with real coefficients is called a *polynomial ring* and denoted by $\mathbb{R}[x_1, \ldots, x_n]$. Let $f_1, \ldots, f_s$ be the polynomials in $\mathbb{R}[x_1, \ldots, x_n]$. A *variety* or an *algebraic set* $V(f_1, \ldots, f_s)$ is the set $\{(a_1, \ldots, a_n) \in \mathbb{R}^n : f_i(a_1, \ldots, a_n) = 0 \text{ for all } 1 \le i \le s\}$. Thus, an algebraic set is the set of all solutions of a system of polynomial equations.

A subset $V$ of $\mathbb{R}^n$ is called a *semi-algebraic set* if $V = \cup_{i=1}^s \cap_{j=1}^{r_i} \{x \in \mathbb{R}^n : P_{i,j}(x) \Leftrightarrow_{ij} 0\}$ where $P_{ij}$ are polynomials in $\mathbb{R}[x_1, \ldots, x_n]$ and $\Leftrightarrow_{ij}$ is one of the comparison operators $\{<, =, >\}$. Informally, a semi-algebraic set is a set that can be described by a finite number of polynomial equalities and inequalities.

An *ideal* $I$ is a subset of a ring, which is closed under addition and multiplication by any polynomial in the ring. The ideal generated by a set of polynomials $g_1, \ldots, g_n$ is the set of polynomials $h$ that can be written as $h = \sum_{i=1}^n f_i g_i$ where $f_i$ are polynomials in the ring and is denoted by $\langle g_1, \ldots, g_n \rangle$. The sum of two ideals $I$ and $J$ is the set $I + J = \{f + g : f \in I, g \in J\}$ and it holds that if $I = \langle f_1, \ldots, f_r \rangle$ and $J = \langle g_1, \ldots, g_s \rangle$, then $I + J = \langle f_1, \ldots, f_r, g_1, \ldots, g_s \rangle$.

## 2.3 Problem

We now define the *implicitization* problem for a set of interventional distributions. We explain what the polynomial constraints computed by the implicitization problem mean algebraically.

Let $\boldsymbol{P}_{intv}$ denote a set of interventional distributions. For example, $\boldsymbol{P}_{intv} = \{P(v_1, v_2), P_{V_1=1}(V_1 = 1, v_2)\}$ contains a non-experimental distribution $P(v_1, v_2)$ and an interventional distribution $P_{V_1=1}(V_1 = 1, v_2)$ where the treatment variable $V_1$ is fixed to 1. We will regard $P(v)$ to be a special interventional distribution where $T = \emptyset$ allowing it to be in $\boldsymbol{P}_{intv}$. Let $\boldsymbol{P}_*$ denote the set of all interventional distributions $\boldsymbol{P}_* = \{P_t(v) | T \subset V, t \in Dm(T), v \in Dm(V), v \text{ is consistent with } t\}$ where $Dm(T)$ represents the domain of $T$. For example, let $V = \{V_1, V_2\}$ where both variables are binary, then $\boldsymbol{P}_* = \{P(v_1, v_2), P_{V_1=1}(V_1 = 1, v_2), P_{V_1=2}(V_1 = 2, v_2), P_{V_2=1}(v_1, V_2 = 1), P_{V_2=2}(v_1, V_2 = 2)\}$.

We can describe $\boldsymbol{P}_{intv}$ in terms of a polynomial mapping from a set of parameters to the distributions as follows.

First, consider a causal BN $G$ without hidden variables. Let $V_1, \ldots, V_n$ be the vertices of $G$. We denote the joint space parameter defining $P_t(v)$ for $v$ consistent with $t$ by $p_v^t$ and the model parameter defining $P(v_i|pa_i)$ by $q_{v_i pa_i}^i$. The model parameters are subjected to the linear relations $\sum_{v_i} q_{v_i pa_i}^i = 1$. Thus, we have introduced $(d_i - 1) \prod_{\{j|V_j \in PA_i\}} d_j$ model parameters for the vertex $V_i$ where $d_i = |Dm(V_i)|$. Let $J_{\boldsymbol{P}_{intv}}$ denote the set of joint space parameters associated with $\boldsymbol{P}_{intv}$ and $M$ denote the set of model parameters. For example, consider a simple causal BN $V_1 \leftarrow V_2$ in which both variables are binary. Let $\boldsymbol{P}_{intv}$ be the set of two distributions $\{P(v_1, v_2), P_{V_1=1}(V_1 = 1, v_2)\}$. Then, $J_{\boldsymbol{P}_{intv}} = \{p_{11}, p_{12}, p_{21}, p_{22}, p_{11}^{V_1=1}, p_{12}^{V_1=1}\}$ and $M = \{q_{11}^1, q_{12}^1, q_1^2\}$. The mapping related to (2) is

$$\phi : \mathbb{R}^M \to \mathbb{R}^{J_{\boldsymbol{P}_{intv}}},$$
$$p_v^t = \prod_{\{i|V_i \notin T\}} q_{v_i pa_i}^i \tag{6}$$

where $\mathbb{R}^M$ and $\mathbb{R}^{J_{\boldsymbol{P}_{intv}}}$ denote the real vector space of dimension $|M|$ and $|J_{\boldsymbol{P}_{intv}}|$ respectively. (6) induces a ring homomorphism

$$\Phi : \mathbb{R}[J_{\boldsymbol{P}_{intv}}] \to \mathbb{R}[M]. \tag{7}$$

Second, consider a causal BN $G$ with hidden variables. Let $\{V_1, \ldots, V_n\}$ and $\{U_1, \ldots, U_{n'}\}$ be sets of observed and hidden variables respectively. We denote the joint space parameters defining $P_t(v)$ for $v$ consistent with $t$ by $p_v^t$ and the model parameters defining $P(v_i|pa_i, u^i)$ and $P(u_j)$ by $q_{v_i pa_i u^i}^i$ and $r_{u_j}^j$ respectively. The joint space parameters and the model parameters form two rings of polynomials $\mathbb{R}[J_{\boldsymbol{P}_{intv}}]$ and $\mathbb{R}[M]$. The mapping related to (4) is

$$\pi : \mathbb{R}^M \to \mathbb{R}^{J_{\boldsymbol{P}_{intv}}},$$
$$p_v^t = \sum_{u_1 \ldots u_{n'}} \prod_{\{i|V_i \notin T\}} q_{v_i pa_i u^i}^i \prod_{j=1}^{n'} r_{u_j}^j. \tag{8}$$

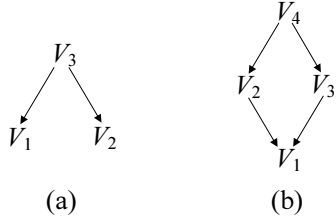

Figure 1: Two causal BNs.

(8) induces a ring homomorphism

$$\Psi : \mathbb{R}[J_{P_{intv}}] \to \mathbb{R}[M]. \tag{9}$$

By Tarski-Seidenberg theorem, the image of $\phi$ (or $\pi$) corresponds to a semi-algebraic set, which can be described by a set of polynomial equalities and inequalities. Finding all of these equalities and inequalities is usually infeasible. In this paper, we choose to find a set of polynomial equalities that define the smallest algebraic set that contains the image of $\phi$ (or $\pi$). These polynomial equalities are a subset of the constraints that describe the image of $\phi$ (or $\pi$) and turn out to be equal to the *kernel* of the ring homomorphism $\Phi$ (or $\Psi$). The *kernel* of $\Phi$, denoted by $\ker(\Phi)$ is the ideal consisting of all polynomials $f$ in $\mathbb{R}[J_{P_{intv}}]$ such that $\Phi(f) = 0$. Thus, the vanishing of the polynomial equalities in $\ker(\Phi)$ and $\ker(\Psi)$ is a necessary condition that there exist the model parameters in (6) and (8) respectively. The process of computing $\ker(\Phi)$ is called *implicitization*.

Our goal is to compute and analyze the kernels for causal BNs with or without hidden variables.

## 3 Causal Bayesian Network with No Hidden Variables

Consider a causal BN $G$ and a set of interventional distributions $P_{intv}$. If checking whether each $P_t(v) \in P_{intv}$ factors as in (2) is the only goal, it is not necessary to solve the implicitization problem since you can use the constraints (2) given by the definition or the constraints given in [Pearl, 2000, pp.23-4]. However, we study the implicitization problem for a set of interventional distributions associated with a causal BN without hidden variables, since we expect that the structure of the constraints for a causal BN without hidden variables may reveal some syntactic structure of the constraints for a causal BN with hidden variables. For non-experimental distribution, [Garcia *et al.*, 2005] showed that the constraints for a BN without hidden variables can help finding the structure of the constraints for a BN with hidden variables.

Since the computation of the constraints for causal BNs without hidden variables is relatively easy, we will focus on the analysis of the computed constraints. In this section, we give a preliminary result on the algebraic structure of the constraints for a set of interventional distributions associated with causal BNs without hidden variables. The problem of characterizing the structure of the constraints for arbitrary set of interventional distributions is still open. We show a few cases in which the constraints can be nicely described by a simple set of polynomials.

### 3.1 One Interventional Distribution

Suppose $P_{intv}$ contains only one interventional distribution $P_t(v)$. For non-experimental distribution $P(v)$, [Garcia *et al.*, 2005] showed that

$$\ker(\Phi) = (I_{\text{local}(G)} : \mathbf{p}^\infty) + \langle \sum_v p_v - 1 \rangle \tag{10}$$

where $I_{\text{local}(G)}$ is the ideal associated to the local Markov property on a BN $G$ and $\mathbf{p}$ is the product of all linear forms $p_{+\ldots+v_{r+1}\ldots v_n} = \sum_{v_1,\ldots,v_r} p_{v_1\ldots v_r v_{r+1}\ldots v_n}$ and $I : f^\infty = \{g \in \mathbb{R}[J_{\{P(v)\}}] \mid gf^N \in I, \text{ for some } N\}$ denotes the *saturation* of $I$ by $f$.

The local Markov property on $G$ is the set of independence statements

$$\text{local}(G) = \{V_i \perp\!\!\!\perp \text{ND}(V_i) | \text{PA}(V_i) : i = 1, \ldots, n\} \tag{11}$$

where $\text{ND}(V_i)$ denotes the set of nondescendents of $V_i$ in $G$ and $\text{PA}(V_i)$ denotes the set of parents of $V_i$ in $G$.

For example, consider the causal BN $G$ in Figure 1 (a). Assume that all variables are binary. The local Markov property on $G$ has only one element $V_1 \perp\!\!\!\perp V_2 \mid V_3$. The constraints induced by an independence statement, $A \perp\!\!\!\perp B \mid C$ are given by the vanishing of the polynomials

$$P(A = a, B = b, C = c)P(A = a', B = b', C = c)$$
$$- P(A = a', B = b, C = c)P(A = a, B = b', C = c) \tag{12}$$

for all $a, a', b, b', c$. Thus, the ideal $I_{\text{local}(G)}$ associated with the local Markov property on $G$ is

$$I_{\text{local}(G)} = \langle p_{111}p_{221} - p_{121}p_{211}, p_{112}p_{222} - p_{122}p_{212} \rangle. \tag{13}$$

For this particular BN $G$, it turns out that

$$I_{\text{local}(G)} : \mathbf{p}^\infty = I_{\text{local}(G)} : (p_{111} \ldots p_{222}p_{+11} \ldots p_{+22}p_{++1}p_{++2})^\infty$$
$$= I_{\text{local}(G)}. \tag{14}$$

From (10), it follows that

$$\ker(\Phi) = I_{\text{local}(G)} + \langle \sum_v p_v - 1 \rangle. \tag{15}$$

In general, however, $\ker(\Phi)$ does not coincide with $I_{\text{local}(G)}$. For example, $I_{\text{local}(G)} : \mathbf{p}^\infty$ for the causal BN $G$ in Figure 1 (b) includes 16 additional generators other than $I_{\text{local}(G)}$.

The above result can be applied to an arbitrary interventional distribution $P_t(v)$. We see that the mapping in (6) defined for $P_t(v)$ and $G$ is equivalent to the mapping defined for $P(v \setminus t)$ and $G(V \setminus T)$ where $G(C)$ denotes the subgraph of $G$ composed only of the variables in $C$. Thus, the following holds.

**Proposition 1** *Let $\Phi$ be a ring homomorphism*

$$\Phi : \mathbb{R}[J_{\{P_t(v)\}}] \to \mathbb{R}[M] \qquad (16)$$

*induced by (6). Then, we have*

$$\ker(\Phi) = (I_{local(G(V \setminus T))} : \mathbf{p}^\infty) + \langle \sum_{v \setminus t} p_v^t - 1 \rangle \qquad (17)$$

where $\mathbf{p}$ is the product of all linear forms $p_{+\ldots+v_{i_{r+1}}\ldots v_{i_k}}$ when $V \setminus T = \{V_{i_1}, \ldots, V_{i_k}\}, V_{i_1} > \ldots > V_{i_k}$.

### 3.2 All Interventional Distributions

Consider the set of all interventional distributions $\boldsymbol{P}_*$. For any joint space parameter $p_v^t$, we have

$$p_v^t = \prod_{\{i|V_i \notin T\}} q_{v_i pa_i}^i = \prod_{\{i|V_i \notin T\}} p_{v_i}^{v \setminus v_i}. \qquad (18)$$

Thus, every joint space parameter can be written as the product of some other joint space parameters. Then,

$$\ker(\Phi) = \langle p_v^t - \prod_{\{i|V_i \notin T\}} p_{v_i}^{v \setminus v_i} : \forall v, t \rangle. \qquad (19)$$

### 3.3 Two Interventional Distributions

Consider the case in which $\boldsymbol{P}_{intv}$ has two distributions. We show some cases in which $\ker(\Phi)$ can be described by a simple set of polynomials.

Consider the causal BN $G$ in Figure 1 (a) where all variables are binary. Suppose $\boldsymbol{P}_{intv} = \{P(v), P_{V_1=1}(v)\}$. We have the following relation between $p_{1v_2v_3}^{V_1=1}$ and $p_v$. For any $v_2$ and $v_3$,

$$p_{1v_2v_3}^{V_1=1} = \sum_{v_1} p_{v_1 v_2 v_3}. \qquad (20)$$

Let $\Phi$ denote a ring homomorphism

$$\Phi : \mathbb{R}[J_{\{P(v),P_{V_1=1}(v_2,v_3)\}}] \to \mathbb{R}[M]. \qquad (21)$$

Since the joint space parameter $p_{1v_2v_3}^{V_1=1}$ for any $v_2$ and $v_3$ is a polynomial function of some of joint space parameters $p_v$, we have

$$\ker(\Phi) = \ker(\Phi') + \langle p_{1v_2v_3}^{V_1=1} - \sum_{v_1} p_{v_1 v_2 v_3} : \forall v_2, v_3 \rangle \qquad (22)$$

where $\Phi'$ denotes the ring homomorphism

$$\Phi' : \mathbb{R}[J_{\{P(v)\}}] \to \mathbb{R}[M]. \qquad (23)$$

From (15), it follows that

$$\ker(\Phi) = I_{local(G)} + \langle \sum_v p_v - 1 \rangle + \langle p_{1v_2v_3}^{V_1=1} - \sum_{v_1} p_{v_1 v_2 v_3} : \forall v_2, v_3 \rangle. \qquad (24)$$

Note that the equation in (20) holds because the set $\{V_2, V_3\}$ contains its own ancestors in $G$. We have the following proposition.

**Proposition 2** *Suppose $\boldsymbol{P}_{intv} = \{P(v), P_t(v)\}$. Let $\Phi$ and $\Phi'$ be ring homomorphisms*

$$\Phi : \mathbb{R}[J_{\{P(v),P_t(v)\}}] \to \mathbb{R}[M], \ \Phi' : \mathbb{R}[J_{\{P(v)\}}] \to \mathbb{R}[M]. \qquad (25)$$

*If $V \setminus T$ contains its own ancestors in $G$, we have*

$$\ker(\Phi) = \ker(\Phi') + \langle p_v^t - \sum_t p_v : \forall (v \setminus t) \rangle. \qquad (26)$$

The relationship between two distributions in the above proposition is the result of Lemma 3 in Section 4.

Now consider the causal BN $G$ in Figure 1 (a) and suppose that $\boldsymbol{P}_{intv} = \{P(v), P_{V_3=1}(v)\}$. In this case, $P_{V_3=1}(v)$ cannot be represented as a polynomial function of $P(v)$. However, we can describe the generators of $\ker(\Phi)$ as follows. Given an instantiation of all the variables $v$ and an instantiation of treatment variables $t$, let $V_{cons} = \{V_i \in V \setminus T \mid v_i pa_i$ in $v$ is consistent with $t\}$ and $\text{cons}(v,t)$ denote the instantiation of $V$ obtained by replacing the inconsistent variables in $v$ with the values of $t$. For example, for $G$ in Figure 1 (a), if $v = (V_1 = 1, V_2 = 1, V_3 = 1)$ and $t = (V_2 = 2)$, then $V_{cons} = \{V_1, V_3\}$ and $\text{cons}(v,t) = (V_1 = 1, V_2 = 2, V_3 = 1)$. We have the following lemma.

**Lemma 1** *Suppose $\boldsymbol{P}_{intv} = \{P(v), P_t(v)\}$. Let $\Phi$, $\Phi'$ and $\Phi''$ be ring homomorphisms*

$$\Phi : \mathbb{R}[J_{\{P(v),P_t(v)\}}] \to \mathbb{R}[M], \ \Phi' : \mathbb{R}[J_{\{P(v)\}}] \to \mathbb{R}[M]$$
$$\Phi'' : \mathbb{R}[J_{\{P_t(v)\}}] \to \mathbb{R}[M]. \qquad (27)$$

*If for any two vertices $V_i$ and $V_j$ in $V \setminus T$, $V_i$ is neither $V_j$'s ancestor nor its descendent, then*

*(i) there exist two disjoint subsets $W_1 = \{A_1, \ldots, A_i\}$ and $W_2 = \{C_1, \ldots, C_k\}$ of $T$ such that*

$$A_1 > \ldots > A_i > B_1 > \ldots > B_j > C_1 > \ldots > C_k \qquad (28)$$

*is a consistent topological ordering of variables in $G$ where $V \setminus T = \{B_1, \ldots, B_j\}$ and*

*(ii)*

$$\ker(\Phi) = \ker(\Phi') + \ker(\Phi'')$$
$$+ \langle f(v,t) \sum_{w_1,v_{cons}} p_v - \sum_{w_1} p_v : \forall v \rangle \qquad (29)$$

*where*

$$f(v,t) = \prod_{\{i|V_i \in V_{cons}\}} \sum_{v_{cons} \setminus v_i} p_{cons(v,t)}^t. \qquad (30)$$

See the Appendix for the proof.

We can use Lemma 1 to compute ker(Φ) for the causal BN $G$ in Figure 1 (a) and $\boldsymbol{P}_{intv} = \{P(v), P_{V_3=1}(v)\}$ since $V_1$ is neither $V_2$'s ancestor nor its descendent. It turns out that

$$\begin{aligned}\ker(\Phi) &= \ker(\Phi') + \ker(\Phi'') + \langle p_{v_1v_21}^{V_3=1} \sum_{v_1,v_2} p_{v_1v_21} - p_{v_1v_21} : \forall v_1, v_2 \rangle \\ &= I_{local(G)} + \langle \sum_v p_v - 1 \rangle + I_{local(G(\{V_1,V_2\}))} + \langle \sum_{v_1,v_2} p_v^{V_3=1} - 1 \rangle \\ &\quad + \langle p_{v_1v_21}^{V_3=1} \sum_{v_1,v_2} p_{v_1v_21} - p_{v_1v_21} : \forall v_1, v_2 \rangle. \end{aligned} \quad (31)$$

## 4 Causal Bayesian Network with Hidden Variables

Solving the implicitization problem for a causal BN with hidden variables has a high computational demand. The implicitization problem can be solved by computing a certain Groebner basis and it is known that computing a Groebner basis has the generic complexity $m^{O(1)}g^{O(N)}$ where $m$ is the number of equations, $g$ is the degree of the polynomials and $N$ is the number of variables. In our implicitization problems, $N$ is the sum of the number of joint space parameters and model parameters. Consider the implicitization for non-experimental distribution. The number of joint space parameters for non-experimental distribution is $d_1 \ldots d_n$. Solving the implicitization problem becomes intractable as the number of vertices in the causal BN and the domains of variables increase. Now consider the cases in which we have a set of interventional distributions. The number of joint space parameters for $\boldsymbol{P}_*$ is $d_1 \ldots d_n (d_1 \ldots d_n - 1)$. This greatly increases the complexity of the already hard problem. In this section, we show two methods to reduce the complexity of our implicitization problem.

### 4.1 Two-step Method

[Garcia *et al.*, 2005] proposed a two-step method to compute ker(Ψ) for a BN with hidden variables and non-experimental distribution. It is known that this method usually works faster than direct implicitization. We apply it to our problem in which we have a set of interventional distributions.

Suppose we have a causal BN $G$ with $n$ observed variables $V_1, \ldots, V_n$ and $n'$ unobserved variables $U_1, \ldots, U_{n'}$ and a set of interventional distributions $\boldsymbol{P}_{intv}$ for $G$. Let Ψ be the ring homomorphism defined in (9). We denote $\boldsymbol{P}_{intv}^U$ be the set of joint distributions assuming that all $U_1, \ldots, U_{n'}$ are observed

$$\boldsymbol{P}_{intv}^U = \{P_t(vu) | P_t(v) \in \boldsymbol{P}_{intv}\}. \quad (32)$$

Let Φ denote the ring homomorphism

$$\Phi : \mathbb{R}[J_{\boldsymbol{P}_{intv}^U}] \to \mathbb{R}[M] \quad (33)$$

induced by the mapping

$$p_{vu}^t = \prod_{\{i|V_i \notin T\}} q_{v_i pa_i u^i}^i \prod_{j=1}^{n'} r_{u_j}^j. \quad (34)$$

For the non-experimental distribution $P(v)$, [Garcia *et al.*, 2005] showed that

$$\ker(\Psi) = \ker(\Phi) \cap \mathbb{R}[J_{\{P(v)\}}]. \quad (35)$$

It can be naturally extended to the case of arbitrary $\boldsymbol{P}_{intv}$. We have

$$\ker(\Psi) = \ker(\Phi) \cap \mathbb{R}[J_{\boldsymbol{P}_{intv}}]. \quad (36)$$

Following [Garcia *et al.*, 2005], ker(Ψ) can be computed in two steps. First, we compute ker(Φ) corresponding to the case where all variables are assumed to be observed. Then we compute the subset of ker(Φ) that corresponds to the polynomial constraints on observable distributions. We have implemented our method using a computer algebra system, Singular [Greuel *et al.*, 2005].

### 4.2 Reducing the Implicitization Problem Using Known Constraints

We can reduce the complexity of the implicitization problem by using some known constraints among interventional distributions. Given the set of joint space parameters $J_{\boldsymbol{P}_{intv}}$, suppose that we have some known constraints among $J_{\boldsymbol{P}_{intv}}$ stating that a joint space parameter $p_v^t$ can be represented as a polynomial function of some other joint space parameters in $J_{\boldsymbol{P}_{intv}} \setminus p_v^t$. Then, the relation reduces the implicitization problem as follows. Let $f$ be a polynomial function such that

$$p_v^t = f(J_{\boldsymbol{P}_{intv}} \setminus p_v^t) \quad (37)$$

and let Ψ and Ψ' be two ring homomorphisms

$$\Psi : \mathbb{R}[J_{\boldsymbol{P}_{intv}}] \to \mathbb{R}[M], \ \Psi' : \mathbb{R}[J_{\boldsymbol{P}_{intv}} \setminus p_v^t] \to \mathbb{R}[M]. \quad (38)$$

Then, we have

$$\ker(\Psi) = \ker(\Psi') + \langle p_v^t - f(J_{\boldsymbol{P}_{intv}} \setminus p_v^t) \rangle. \quad (39)$$

This suggests that the more we find such relations among parameters, the more we can reduce the implicitization problem. The following two lemmas provide a class of such relations.

A *c-component* is a maximal set of vertices such that any two vertices in the set are connected by a path on which every edge is of the form ←-- $U$ --→ where $U$ is a hidden variable. A set $A \subseteq V$ is called an *ancestral set* if it contains its own observed ancestors.

```
procedure PolyRelations(G, J_{P_{intv}})
INPUT: a causal BN G, joint space parameters J_{P_{intv}} associ-
ated with a set of interventional distributions P_{intv}
OUTPUT: a subset J'_{P_{intv}} ⊆ J_{P_{intv}} of joint space parameters and
the ideal I containing polynomial relations among the joint
space parameters
Initialization:
I ← ∅
J'_{P_{intv}} ← J_{P_{intv}}
Step 1:
For each p_v^t ∈ J'_{P_{intv}}
  Let H_1, ..., H_l be the c-components in the subgraph
  G(V \ T).
  I ← I + ⟨p_v^t − ∏_i p_v^{v\h_i}⟩
  J'_{P_{intv}} ← J'_{P_{intv}} \ p_v^t
Step 2:
For each p_v^t ∈ J'_{P_{intv}}
  If there is a joint space parameter p_v^c that satisfies
  (i) C ⊆ T ⊆ V
  (ii) V \ T is an ancestral set in G(V \ C)
  then
    I ← I + ⟨p_v^t − ∑_{t\c} p_v^c⟩
    J'_{P_{intv}} ← J'_{P_{intv}} \ p_v^t
```

Figure 2: A Procedure for Listing Polynomial Relations among Interventional Distributions

**Lemma 2** [Tian and Pearl, 2002] *Let $T \subseteq V$ and assume that $V \setminus T$ is partitioned into c-components $H_1, \ldots, H_l$ in the subgraph $G(V \setminus T)$. Then we have*

$$P_t(v) = \prod_i P_{v \setminus h_i}(v). \tag{40}$$

**Lemma 3** [Tian and Pearl, 2002] *Let $C \subseteq T \subseteq V$. If $V \setminus T$ is an ancestral set in $G(V \setminus C)$, then*

$$P_t(v) = \sum_{t \setminus c} P_c(v). \tag{41}$$

We give a procedure in Figure 2 that lists a set of polynomial relations among $P_{intv}$ based on these two lemmas. Given a set of joint space parameters $J_{P_{intv}}$, it outputs a subset $J'_{P_{intv}}$ of $J_{P_{intv}}$ which contains the joint space parameters that cannot be represented as a polynomial function of other joint space parameters, and the ideal $I$ generated by all the relations found by Lemma 2 and Lemma 3. In Step 1, we look for the parameters that can be represented as the product of other parameters using Lemma 2. In Step 2, we find the parameters that can be represented as the sum of other parameters using Lemma 3. We have the following proposition.

**Proposition 3** *Given a set of interventional distributions $P_{intv}$, a causal BN $G$ with hidden variables and a ring homomorphism $\Psi$ defined in (9), let $J'_{P_{intv}}$ and $I$ be the results*

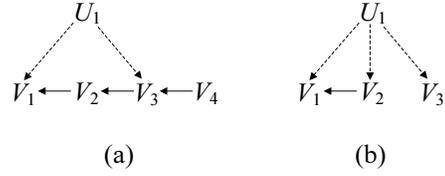

Figure 3: Two causal BNs with one hidden variable.

*computed by* **PolyRelations**. *Then,*

$$\ker(\Psi) = \ker(\Psi') + I \tag{42}$$

*where $\Psi'$ is a ring homomorphism*

$$\Psi' : \mathbb{R}[J'_{P_{intv}}] \to \mathbb{R}[M]. \tag{43}$$

To illustrate the procedure, consider a causal BN $G$ with four observed variables $V_1, V_2, V_3, V_4$ and one hidden variable $U_1$ in Figure 3 (a). We will compute $\ker(\Psi)$ for the set of all interventional distributions $P_*$ using **PolyRelations**. In Step 1, we find that most of joint space parameters can be represented as the product of other parameters. For example, we have

$$p_v^{v_1} = p_v^{v_1 v_3 v_4} p_v^{v_1 v_2 v_4} p_v^{v_1 v_2 v_3} \tag{44}$$

since $V \setminus V_1 = \{V_2, V_3, V_4\}$ is partitioned into three c-components $\{V_2\}, \{V_3\}$ and $\{V_4\}$. Also,

$$p_v^{v_2} = p_v^{v_2 v_4} p_v^{v_1 v_2 v_3} \tag{45}$$

since $V \setminus V_2 = \{V_1, V_3, V_4\}$ is partitioned into two c-components $\{V_1, V_3\}$ and $\{V_4\}$. The only joint space parameters that do not decompose in Step 1 are

$$p_v^{v_2 v_4}, p_v^{v_1 v_3 v_4}, p_v^{v_1 v_2 v_3}, p_v^{v_2 v_3 v_4} \text{ and } p_v^{v_1 v_2 v_4}. \tag{46}$$

Thus, after Step 1 we have

$$J'_{P_{intv}} = J_{\{P_{v_2 v_4}(v), P_{v_1 v_3 v_4}(v), P_{v_1 v_2 v_3}(v), P_{v_2 v_3 v_4}(v), P_{v_1 v_2 v_4}(v)\}}. \tag{47}$$

In Step 2, we find that

$$p_v^{v_2 v_3 v_4} = \sum_{v_3} p_v^{v_2 v_4} \text{ and } p_v^{v_1 v_2 v_4} = \sum_{v_1} p_v^{v_2 v_4} \tag{48}$$

since $V \setminus \{V_2, V_3, V_4\} = \{V_1\}$ and $V \setminus \{V_1, V_2, V_4\} = \{V_3\}$ are ancestral sets in $G(V \setminus \{V_2, V_4\}) = G(\{V_1, V_3\})$. After Step 2, we have

$$J'_{P_{intv}} = J_{\{P_{v_2 v_4}(v), P_{v_1 v_3 v_4}(v), P_{v_1 v_2 v_3}(v)\}} \tag{49}$$

and $I$ is generated by all the relations found in Step 1 and 2. Finally, we have

$$\ker(\Psi) = \ker(\Psi') + I \tag{50}$$

where $\Psi'$ is the ring homomorphism

$$\Psi': \mathbb{R}[J'_{P_{intv}}] \to \mathbb{R}[M]. \tag{51}$$

Moreover, we find that $\ker(\Psi')$ can be represented as $\ker(\Psi_1) + \ker(\Psi_2) + \ker(\Psi_3)$ where

$$\Psi_1: \mathbb{R}[J_{\{P_{v_2 v_4}(v)\}}] \to \mathbb{R}[M], \; \Psi_2: \mathbb{R}[J_{\{P_{v_1 v_3 v_4}(v)\}}] \to \mathbb{R}[M]$$
$$\Psi_3: \mathbb{R}[J_{P_{\{v_1 v_2 v_3}(v)\}}] \to \mathbb{R}[M] \tag{52}$$

since the mappings inducing $\Psi_1$, $\Psi_2$ and $\Psi_3$ do not share model parameters. This gives

$$\ker(\Psi) = \ker(\Psi_1) + \ker(\Psi_2) + \ker(\Psi_3) + I. \tag{53}$$

Compared to the original implicitization problem of computing $\ker(\Psi)$ involving 240 joint space parameters which is intractable, we now have three small implicitization problems. Computing $\ker(\Psi_1)$ involves 12 joint space parameters and each of the computation of $\ker(\Psi_2)$ and $\ker(\Psi_3)$ involves 2 joint space parameters. The reduced problem can be solved easily.

Note that $J'_{P_{intv}}$ computed by **PolyRelations** in the above example contains only the joint space parameters related to c-components in $G$. This holds generally for $G$ in which the subgraph $G(C)$ for each c-component $C$ of $G$ has no edges.

**Proposition 4** *Let $C_1, \ldots, C_l$ be c-components of a causal BN $G$. If every subgraph $G(C_i)$ has no edges, then*

$$\ker(\Psi) = \ker(\Psi_1) + \ldots + \ker(\Psi_l) + I \tag{54}$$

*where*

$$\Psi_i: \mathbb{R}[J_{\{P_{v \setminus c_i}(v)\}}] \to \mathbb{R}[M] \tag{55}$$

*and $I$ is the ideal computed by the procedure* **PolyRelations**.

The implicitization problem for a large causal BN $G$ is computationally feasible if $G$ has the structure described in Proposition 4 and the size of each c-component in $G$ is small. Our method becomes infeasible as the size of each c-component grows.

In general, there may be some constraints that are not included in the constraints for each c-component and cannot be found by Lemma 2 and 3. For example, for the causal BN $G$ in Figure 3 (b), we find the following constraint by the method in Section 4.1 using the Singular system:

$$p_{222} p_{122}^{V_2=2} p_{211}^{V_2=1} + p_{222} p_{122}^{V_2=2} p_{212}^{V_2=1} + p_{212} p_{122}^{V_2=2} p_{221}^{V_2=2}$$
$$+ p_{122} p_{212}^{V_2=1} p_{221}^{V_2=2} + p_{222} p_{212}^{V_2=1} p_{221}^{V_2=2} - p_{122}^{V_2=2} p_{212}^{V_2=1} p_{221}^{V_2=2}$$
$$+ p_{212} p_{122}^{V_2=2} p_{222}^{V_2=2} - p_{122} p_{211}^{V_2=1} p_{222}^{V_2=2} + p_{222} p_{212}^{V_2=1} p_{222}^{V_2=2}$$
$$- p_{122}^{V_2=2} p_{212}^{V_2=1} p_{222}^{V_2=2} + p_{221}^{V_2=2} p_{222}^{V_2=2} - p_{212}^{V_2=1} p_{221}^{V_2=2} p_{222}^{V_2=2}$$
$$+ p_{212} p_{222}^{V_2=2} p_{222}^{V_2=2} - p_{212} p_{222}^{V_2=1} p_{222}^{V_2=2} - p_{222} p_{212}^{V_2=1} - p_{212} p_{222}^{V_2=2}$$
$$+ p_{212}^{V_2=1} p_{222}^{V_2=2} \tag{56}$$

which is in $\ker(\Psi)$ but cannot be induced by Lemma 2 and 3.

## 5 Conclusion and Future Work

We obtain polynomial constraints on the interventional distributions induced by a causal BN with hidden variables, via the implicitization procedure. These constraints constitute a necessary test for a causal model to be compatible with given observational and experimental data. To apply these constraints to finite data in practice, an important future work is to design test statistics for non-independence constraints. Another future work is to study how to use these constraints in the model selection process. We are investigating a model selection method that uses a new goodness-of-fit score based on the geometric distance between data and a model.

We are also working on the general characterization of the constraints computed by implicitization for causal BNs without hidden variables, which will be helpful in finding the algebraic structure of the constraints implied by causal BNs with hidden variables which typically have complicated structures.

### Acknowledgments

This research was partly supported by NSF grant IIS-0347846.

### Appendix : Proof of Lemma 1

We define the ideal $I$ associated with $\Phi$.

$$I = \langle p_v - \prod_i q^i_{v_i p a_i} : \forall v \rangle + \langle p^t_v - \prod_{\{i | V_i \notin T\}} q^i_{v_i p a_i} : \forall (v \setminus t) \rangle. \tag{57}$$

The elimination ideal $I \cap \mathbb{R}[J_{\{P(v), P_t(v)\}}]$ is equivalent to $\ker(\Phi)$. The idea is that we can represent $I$ as the sum of three ideal $I_1$, $I_2$ and $I_3$ such that the model parameters in $I_1$ and those in $I_2$ are disjoint and no model parameter appears in $I_3$ and thus

$$\ker(\Phi) = I \cap \mathbb{R}[J_{\{P(v), P_t(v)\}}]$$
$$= I_1 \cap \mathbb{R}[J_{\{P(v)\}}] + I_2 \cap \mathbb{R}[J_{\{P_t(v)\}}] + I_3$$
$$= \ker(\Phi') + \ker(\Phi'') + I_3. \tag{58}$$

Let $I_1 = \langle p_v - \prod_i q^i_{v_i p a_i} : \forall v \rangle$ and $I_2 = \langle p^t_v - \prod_{\{i | V_i \notin T\}} q^i_{v_i p a_i} : \forall (v \setminus t) \rangle$. We will replace each generator in $I_1$ with two other polynomials and add one polynomial to $I_3$ which is initially empty as follows.

For any polynomial $p_v - \prod_i q^i_{v_i p a_i}$, we have

$$p_v - \prod_i q^i_{v_i p a_i} \tag{59}$$
$$= p_v - \Big(\prod_{\{i | V_i \in W_1\}} q^i_{v_i p a_i}\Big)\Big(\prod_{\{i | V_i \in V \setminus T\}} q^i_{v_i p a_i}\Big)\Big(\prod_{\{i | V_i \in W_2\}} q^i_{v_i p a_i}\Big)$$
$$= p_v - \Big(\prod_{\{i | V_i \in W_1\}} q^i_{v_i p a_i}\Big)\Big(\sum_{w_1} p_v\Big) \tag{60}$$

since

$$\sum_{w_1} p_v - \Big(\prod_{\{i | V_i \in V \setminus T\}} q^i_{v_i p a_i}\Big)\Big(\prod_{\{i | V_i \in W_2\}} q^i_{v_i p a_i}\Big)$$

is in $I$. Also,

$$\sum_{w_1} p_v - \Big(\prod_{\{i|V_i \in V \setminus T\}} q^i_{v_i pa_i}\Big)\Big(\prod_{\{i|V_i \in W_2\}} q^i_{v_i pa_i}\Big)$$
$$= \sum_{w_1} p_v - \Big(\prod_{\{i|V_i \in V_{cons}\}} q^i_{v_i pa_i}\Big)\Big(\prod_{\{i|V_i \in (V \setminus T) \setminus V_{cons}\}} q^i_{v_i pa_i}\Big)$$
$$\Big(\prod_{\{i|V_i \in W_2\}} q^i_{v_i pa_i}\Big)$$

From the property that any two vertices $V_i$ and $V_j$ in $V \setminus T$, $V_i$ is neither $V_j$'s ancestor nor its parent, it follows that the polynomial

$$\sum_{w_1, v_{cons}} p_v - \Big(\prod_{\{i|V_i \in (V \setminus T) \setminus V_{cons}\}} q^i_{v_i pa_i}\Big)\Big(\prod_{\{i|V_i \in W_2\}} q^i_{v_i pa_i}\Big) \quad (61)$$

is in $I$. Thus,

$$\sum_{w_1} p_v - \Big(\prod_{\{i|V_i \in V \setminus T\}} q^i_{v_i pa_i}\Big)\Big(\prod_{\{i|V_i \in W_2\}} q^i_{v_i pa_i}\Big)$$
$$= \sum_{w_1} p_v - \Big(\prod_{\{i|V_i \in V_{cons}\}} q^i_{v_i pa_i}\Big)\Big(\sum_{w_1, v_{cons}} p_v\Big)$$
$$= \sum_{w_1} p_v - \Big(\prod_{\{i|V_i \in V_{cons}\}} \sum_{v_{cons} \setminus v_i} p^t_{cons(v,t)}\Big)\Big(\sum_{w_1, v_{cons}} p_v\Big). \quad (62)$$

We replace the polynomial (59) with the polynomials (60) and (61) and add the polynomial (62) to $I_3$. After processing every polynomial in $I_1$, we have three ideal $I_1$, $I_2$ and $I_3$ with the desired property. ∎

## References

[Desjardins, 1999] B. Desjardins. *On the theoretical limits to reliable causal inference*. PhD thesis, University of Pittsburgh, 1999.

[Garcia *et al.*, 2005] L.D. Garcia, M. Stillman, and B. Sturmfels. Algebraic geometry of bayesian networks. *Journal of Symbolic Computation*, 39(3–4):331–355, 2005.

[Garcia, 2004] Luis David Garcia. Algebraic statistics in model selection. In *Proceedings of the 20th Annual Conference on Uncertainty in Artificial Intelligence (UAI-04)*, pages 177–18, Arlington, Virginia, 2004. AUAI Press.

[Geiger and Meek, 1998] Dan Geiger and Christopher Meek. Graphical models and exponential families. In *Proceedings of the Fourteenth Annual Conference on Uncertainty in Artificial Intelligence (UAI–98)*, pages 156–165, San Francisco, CA, 1998. Morgan Kaufmann Publishers.

[Geiger and Meek, 1999] Dan Geiger and Christopher Meek. Quantifier elimination for statistical problems. In *Proceedings of the Fifteenth Annual Conference on Uncertainty in Artificial Intelligence (UAI–99)*, pages 226–235, San Francisco, CA, 1999. Morgan Kaufmann Publishers.

[Greuel *et al.*, 2005] G.-M. Greuel, G. Pfister, and H. Schönemann. SINGULAR 3.0. A Computer Algebra System for Polynomial Computations, Centre for Computer Algebra, University of Kaiserslautern, 2005. http://www.singular.uni-kl.de.

[Heckerman and Shachter, 1995] D. Heckerman and R. Shachter. Decision-theoretic foundations for causal reasoning. *Journal of Artificial Intelligence Research*, 3:405–430, 1995.

[Kang and Tian, 2006] C. Kang and J. Tian. Inequality constraints in causal models with hidden variables. In *Proceedings of the Seventeenth Annual Conference on Uncertainty in Artificial Intelligence (UAI-06)*, pages 233–240, Arlington, Virginia, 2006. AUAI Press.

[Lauritzen, 2000] S. Lauritzen. Graphical models for causal inference. In O.E. Barndorff-Nielsen, D. Cox, and C. Kluppelberg, editors, *Complex Stochastic Systems*, chapter 2, pages 67–112. Chapman and Hall/CRC Press, London/Boca Raton, 2000.

[Pearl, 1995] J. Pearl. Causal diagrams for empirical research. *Biometrika*, 82:669–710, December 1995.

[Pearl, 2000] J. Pearl. *Causality: Models, Reasoning, and Inference*. Cambridge University Press, NY, 2000.

[Riccomagno and Smith, 2003] E. Riccomagno and J.Q. Smith. Non-graphical causality: a generalization of the concept of a total cause. Technical Report No. 394, Department of Statistics, University of Warwick, 2003.

[Riccomagno and Smith, 2004] E. Riccomagno and J.Q. Smith. Identifying a cause in models which are not simple bayesian networks. In *Proceedings of IPMU*, pages 1315–1322, Perugia, 2004.

[Robins and Wasserman, 1997] James M. Robins and Larry A. Wasserman. Estimation of effects of sequential treatments by reparameterizing directed acyclic graphs. In *Proceedings of the Thirteenth Annual Conference on Uncertainty in Artificial Intelligence (UAI–97)*, pages 409–420, San Francisco, CA, 1997. Morgan Kaufmann Publishers.

[Spirtes *et al.*, 2001] P. Spirtes, C. Glymour, and R. Scheines. *Causation, Prediction, and Search*. MIT Press, Cambridge, MA, 2001.

[Tian and Pearl, 2002] J. Tian and J. Pearl. On the testable implications of causal models with hidden variables. In *Proceedings of the Conference on Uncertainty in Artificial Intelligence (UAI)*, 2002.

[Verma and Pearl, 1990] T. Verma and J. Pearl. Equivalence and synthesis of causal models. In P. Bonissone et al., editor, *Uncertainty in Artificial Intelligence 6*, pages 220–227. Elsevier Science, Cambridge, MA, 1990.